\documentclass[final]{cvpr}

\usepackage{times}
\usepackage{epsfig}
\usepackage{graphicx}
\usepackage{amsmath}
\usepackage{amssymb}
\usepackage{mathtools}

\usepackage{makecell}
\usepackage{subcaption}

\usepackage[pagebackref=true,breaklinks=true,colorlinks,bookmarks=false]{hyperref}
\renewcommand{\sectionautorefname}{\S}
\renewcommand{\subsectionautorefname}{\S}
\renewcommand{\subsubsectionautorefname}{\S}

\newcommand{\secmention}{Sec. }
\newcommand{\tabmention}{Tab. }
\newcommand{\figmention}{Fig. }
\newcommand{\eqnmention}{Eqn. }

\DeclarePairedDelimiter\floor{\lfloor}{\rfloor}

\begin{document}

\title{MARNet: Multi-Abstraction Refinement Network for 3D Point Cloud Analysis}

\author{Rahul Chakwate \qquad Arulkumar Subramaniam \qquad Anurag Mittal \\
Indian Institute of Technology Madras \\
{\tt\small ae16b005@smail.iitm.ac.in}
}
\maketitle

\begin{abstract}
\label{abstract}

Representation learning from 3D point clouds is challenging due to their inherent nature of permutation invariance and irregular distribution in space. Existing deep learning methods follow a hierarchical feature extraction paradigm in which high-level abstract features are derived from low-level features. However, they fail to exploit different granularity of information due to the limited interaction between these features. To this end, we propose Multi-Abstraction Refinement Network (MARNet) that ensures an effective exchange of information between multi-level features to gain local and global contextual cues while effectively preserving them till the final layer. We empirically show the effectiveness of MARNet in terms of state-of-the-art results on two challenging tasks: Shape classification and Coarse-to-fine grained semantic segmentation. MARNet significantly improves the classification performance by 2\% over the baseline and outperforms the state-of-the-art methods on semantic segmentation task.

\end{abstract}

\vspace{-0.5cm}
\section{Introduction}
\label{sec:introduction}

{\let\thefootnote\relax\footnote{ Code available at: \href{https://github.com/ruc98/MARNet}{\tt https://github.com/ruc98/MARNet} }}

\begin{figure}
\vspace{-0.5cm}
\begin{center}
    \includegraphics[width=\linewidth]{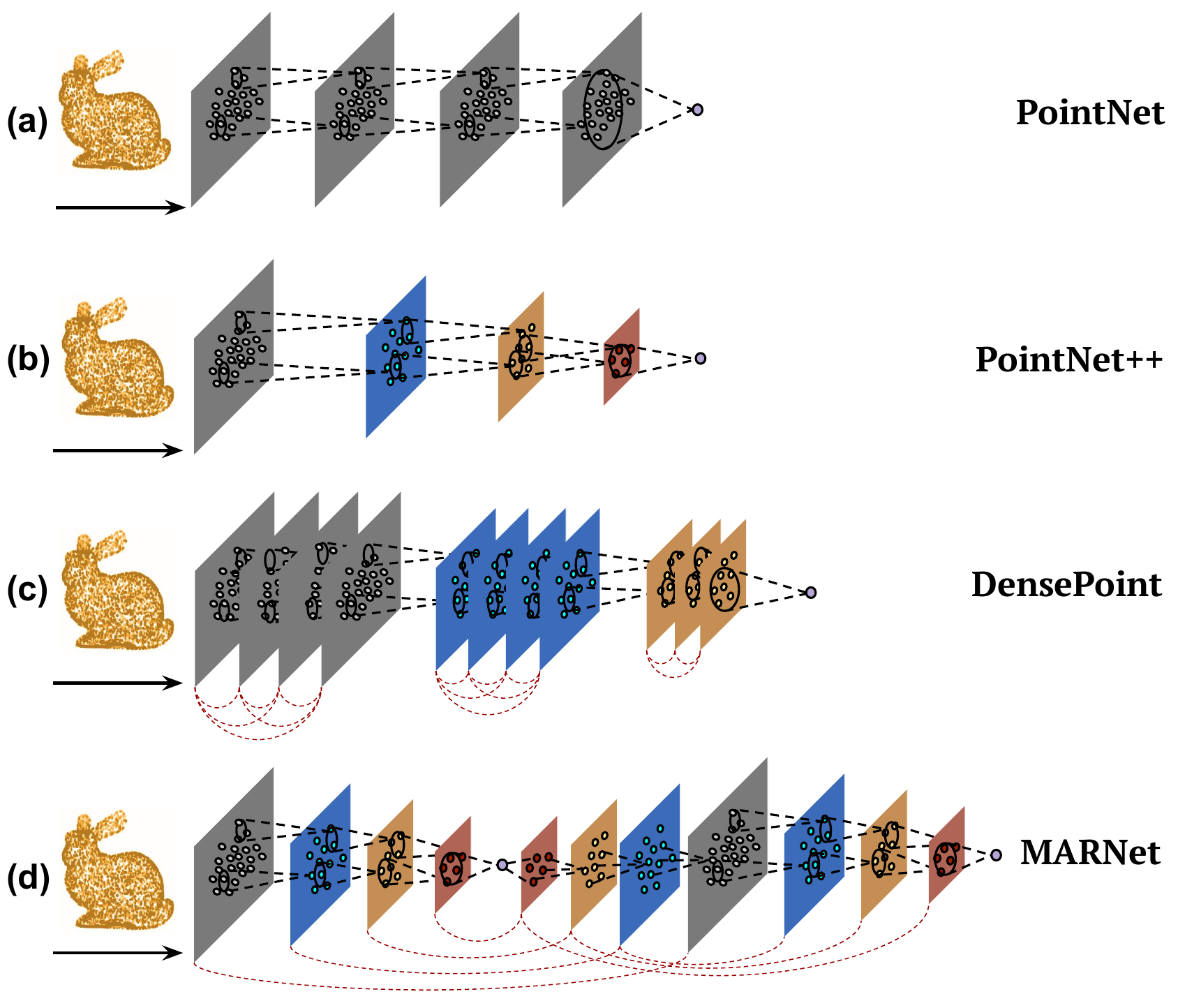}
\end{center}
\vspace{-0.5cm}
\caption{State-of-the-art models for feature extraction from 3D point clouds. (a) PointNet~\cite{pointnet} uses a series of point-wise MLPs followed by max/average pooling, (b) PointNet++~\cite{pointnet2} introduces the notion of hierarchical feature extraction to capture local patterns, (c) DensePoint~\cite{densepoint} uses densely connected blocks to aggregate contextual information, (d) \textbf{MARNet (ours)} proposes a unified architecture to perform hierarchical and multi-level feature aggregation to capture local and global contextual cues.
\label{fig:motivation}}
\vspace{-0.5cm}
\end{figure}

The recent evolution of 3D sensors such as LiDAR and Kinect has boosted the ability to perceive the environment in terms of 3D point clouds. 3D Point cloud analysis has increasingly become ubiquitous in robotic perception~\cite{robotics}, augmented / mixed reality~\cite{ar}  and autonomous driving~\cite{rt3d,kitti,sun2020scalability}. 
Traditional analysis techniques with hand-crafted features are unable to cater to the needs of these modern applications that demand diverse semantic understanding. With a proven track record in terms of state-of-the-art solutions in multiple domains such as image analysis~\cite{alexnet,vgg,resnet,densenet,fishnet} and natural language processing~\cite{all_you_need,besacier2014automatic,otter2018survey}, deep learning pledges a promising alternative to conventional methods. Owing to the remarkable improvements of deep neural network architectures in 2D image analysis, researchers attempt to port and adapt the 2D network architectures to 3D point clouds~\cite{pointnet, pointnet2, pointcnn, densepoint}. 

A natural perspective is to convert irregular point clouds to regular grid voxels~\cite{voxnet,3dshapenets,pointgrid,minkowski} and apply 3D CNNs to extract features from them. However, voxelization may obscure the finer details due to lower grid resolution. On the other hand, the cost of making the voxel grids finer is exponentially high. Another strategy is to project the 3D information on multi-view images~\cite{mvcnn,mhbn,c17_mv,c18_mv}. However, it suffers from the loss of crucial 3D geometric information and thus lack rich contextual features.

A pioneering approach called PointNet~\cite{pointnet} proposes to directly operate on irregular point clouds. It transforms the 3D points using a series of point-wise feature transformations and finally, outputs an aggregated feature vector using a permutation invariant symmetric function. This successful method, however, has a downside that local patterns are not taken into account. To mitigate this issue, PointNet++~\cite{pointnet2} proposes to group the neighboring points in the euclidean space and apply PointNet~\cite{pointnet} locally on each of the group, thus inducing the notion of hierarchical feature extraction. This architectural design resembles the hierarchical feature extraction of CNNs used for 2D image analysis~\cite{alexnet}.

In their solution to incorporate local patterns, PointNet++~\cite{pointnet2} proposes two layers, namely: 1) Multi-Scale Grouping (MSG) layer and 2) Multi-Resolution Grouping (MRG) layer. The MSG layer aggregates point-wise features at different scales (i.e., group the points with multiple radii). Whereas, MRG layer aggregates the point features at different resolutions (i.e., from multiple abstraction layers). Strong empirical performance of~\cite{pointnet2} suggests that both these layers aid in capturing local patterns. However, these layers fail to apprehend dense contextual insights, as there is a lack of feature interaction between the global features in deeper layers and local features in earlier layers. Inspired by architectural improvement in the 2D image analysis domain~\cite{densenet}, DensePoint~\cite{densepoint} tries to diminish this issue by using densely-connected blocks to encourage feature reuse and enhance feature propagation in 3D point clouds. On the downside, it fails to support multi-level feature interaction and lacks in its ability to preserve the features from all stages, as the earlier features are modified during the course of forward propagation.

As a result of the above observations, to achieve optimal performance in the point cloud analysis tasks, we notice that the following requirements are crucial in the network: 1) Multi-scale and multi-resolution aggregation of the features to capture local as well as global patterns, 2) Efficient feature communication between the shallow and deeper features to gain dense-contextual information, 3) Preservation of features at all levels of abstraction for effective feature learning and back-propagation.

To this end,  we propose a novel deep network architecture for 3D point clouds analysis: \textbf{M}ulti-\textbf{A}bstraction \textbf{R}efinement \textbf{N}etwork (MARNet), by carefully designing the network layers to satisfy the above requirements. Precisely, MARNet (\figmention\ref{fig:motivation}) consists of three stages in its network design, namely: 1) Backbone stage, 2) Feature Cross-Referencing (FCR)
stage, and 3) Feature Re-Encoding (FRE) stage. First, the Backbone stage employs a hierarchical multi-scale feature extractor such as PointNet++~\cite{pointnet2}, thus incorporating multi-scale and multi-resolution feature learning. Next, to encourage effective feature propagation between shallow and deep layers, the FCR stage fosters an efficient interaction of the multiple abstraction features from the Backbone stage and allows them to refine each other using a specially devised parameter-less reduction function. Further, FCR and FRE stages are designed to preserve features from all the Backbone levels by incorporating residual connections until the final output layer, thus encouraging unimpeded gradient flow. Our three-stage design of MARNet is inspired by one of the recent deep learning architectural improvements in the 2D image analysis domain called FishNet~\cite{fishnet} that makes use of multi-level feature interaction to achieve state-of-the-art performance in 2D image classification, detection, and segmentation.  

The key contributions of our paper are as follows:
\vspace{-0.2cm}
\begin{itemize}
    \itemsep0em 
    \item We propose a novel unified framework that utilizes the complementary nature of multi-level abstract features and encourages them to interact and refine each other.
    \item We carefully design network layers to preserve point-features of different granularity such that unmodified gradients are passed to earlier layers to overcome vanishing/exploding gradient problem
    \item Through extensive experiments, we verify the effectiveness of MARNet by attaining state-of-the-art results on challenging benchmarks for the tasks: 3D shape classification, coarse-, middle- and fine-grained semantic segmentation.
\end{itemize}

\begin{figure*}[h!]
\begin{center}
\includegraphics[width=\linewidth]{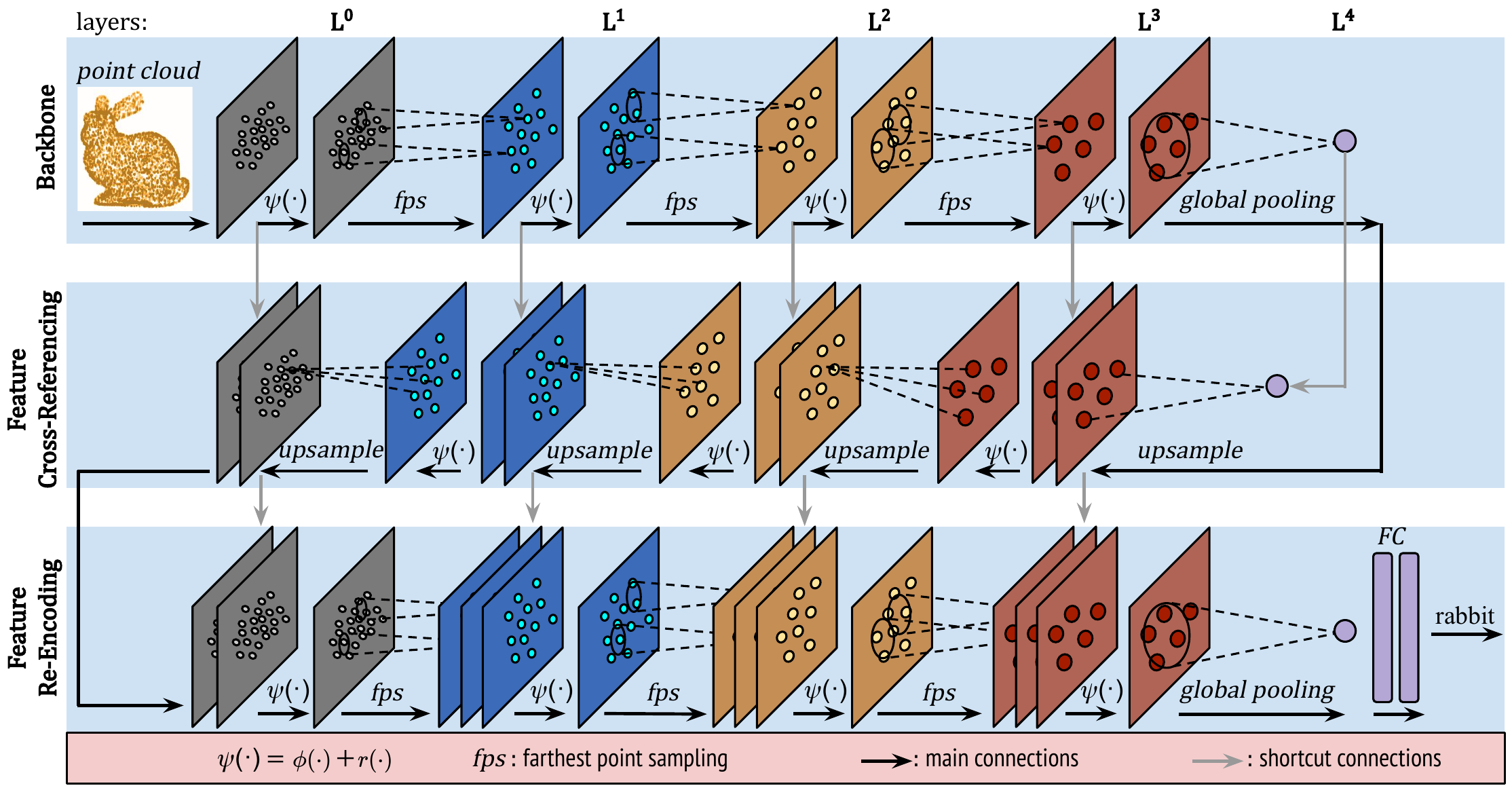}
\end{center}
 \vspace{-0.6cm}
   \caption{The overall network architecture containing three stages of feature extraction/refinement: 1) Backbone stage (top), 2) Feature Cross-Referencing stage (middle) and 3) Feature Re-Encoding stage (bottom). $\psi(\cdot) = $ transformation function $\phi(\cdot)+$ residual function $r(\cdot)$ as mentioned in \secmention  \ref{sec:methodology}. In both Backbone \& FRE stage, $r(\cdot)$ is implemented as identity function, whereas in FCR stage, $r(\cdot)$ is implemented as a reduction function (\secmention \ref{section:fcr}).
\label{fig:overall}}
\vspace{-0.5cm}
\end{figure*}

\vspace{-0.3cm}
\section{Related Work}
\label{sec:related work}
We delineate the existing works on 3D point cloud analysis into the following categories:
\vspace{-0.45cm}
\paragraph{Multi-view projection-based and volumetric methods:}
The multi-view projection-based methods~\cite{mvcnn,mhbn,c17_mv,c18_mv,c19_mv,c20_mv,c21_mv} project a 3D point cloud onto multiple 2D views and apply 2D CNN to extract view based features. These multi-view features are aggregated to get a global representation of the shape. Naively projecting 3D objects onto 2D space leads to loss of valuable geometric information. Other methods~\cite{voxnet,3dshapenets,pointgrid,minkowski,latticenet,mrtnet} project 3D point cloud onto regular 3D grid voxel. However, voxelization leads to quantization loss caused by the low grid resolution of the voxels. The computation increases exponentially with a linear increase in grid resolution. Kd trees~\cite{kdnet} and octree~\cite{octree,octnet,ocnn} based methods alleviate these limitations though they are still dependent on subdivision of volume. On the contrary, our model learns directly from irregular point clouds.

\vspace{-0.45cm}
\paragraph{Point-wise MLP based Methods:}
These types of networks operate directly on each point. PointNet~\cite{pointnet} applies a shared-MLP on each point independently and aggregates the point features using max-pooling to obtain a global representation. However, it fails to capture the local patterns as the features are learned independently for every point. PointNet++~\cite{pointnet2} overcomes this limitation by hierarchically grouping and downsampling the point clouds. Many subsequent networks~\cite{pat,srn,pointwisecnn,geocnn,densepoint} including ours are based on these networks. 

\vspace{-0.45cm}
\paragraph{Convolutional Kernel-Based Methods:}
Kernel-based networks also involve point-wise MLPs. However, they have specialized 3D convolutional kernels that operate either locally or globally on the point clouds. \cite{pointconv, mccnn, spidercnn, pcnn} use existing methods such as Monte Carlo estimation, Taylor expansion, k-nearest neighbors to model the 3D convolution operation. \cite{sphericalcnn,pcnn,sphnet} address the rotation equivariance of point clouds by transforming the problem into polar coordinates. Several methods~\cite{geocnn, pointcnn, interpcnn} cornerstone the point clouds' inherent geometric aspects while~\cite{densepoint, shapecontextnet, rscnn} focus on context aggregation. In contrast, our method focuses on preserving and refining the features obtained from the points and making the best use of the available features.

\vspace{-0.45cm}
\paragraph{Graph Based Methods:}
These networks model point clouds as a graph with each point representing the vertex and the connection between the points as edges.~\cite{eccnet, kcnet, dgcnn, ldgcnn} model the distance between the points as the weights of the edges to capture the local geometry of the points.~\cite{dpam, clusternet} aim at easing the task of point agglomeration into simple steps.~\cite{rgcnn, agcn, hgnn,specgcn} exploit the spectral domains for graph creation.

\vspace{-0.2cm}
\section{Methodology}
\label{sec:methodology}
Our network architecture consists of three stages of feature extraction and refinement, namely: 1) Backbone stage, 2) Feature Cross-Referencing stage (FCR) stage, and 3) Feature Re-Encoding (FRE) stage. 
The proposed network architecture is shown in \figmention \ref{fig:overall}.

The stages of our network are explained in Sec. \ref{section:backbone}, Sec. \ref{section:fcr} and Sec. \ref{section:fre} respectively. 
The unique features of MARNet, such as multi-level feature aggregation and unimpeded backpropagation of gradients, are discussed in \secmention \ref{section:gradients}.

\vspace{-0.1cm}
\subsection{Mathematical Notation}
\label{section:notation}

The different "levels" of abstraction are labeled as ${ L}^i$ where $i\in [0,1,\dots]$ denotes the level index. The collection of 3D points at level ${ L}^i$ is given as  ${P}^i  = [ {p}_1, {p}_2,\dots,{p}_{N^i}]\in \mathbb{R}^{N^i \times 3}$ where $N^i$ is the number of 3D points at level ${L}^i$ and dimensions of 3D points are $[x,y,z]$. 
The collection of point-wise features at level ${ L}^i$ is given by ${F}^i  = [ {f}({p_1}),{f}({p_2}),\dots,{f}({p_{N^i}})]\in \mathbb{R}^{N^i \times D^i}$, where ${f}({p_i}) \in \mathbb{R}^{D^i}$ denotes the point-wise feature vector corresponding to the 3D point ${p}_i$, $D^i$ = feature dimension at level ${ L}^i$.

Any entity $\mathbb{X}$ (such as Level ${ L}^i$, 3D points ${p}^i$, point features ${f}^i$) corresponding to different stages (Backbone, FCE, FRE) are denoted by subscripts: $\mathbb{X}_{bb}$ for Backbone, $\mathbb{X}_{fcr}$ for FCR stage and $\mathbb{X}_{fre}$ for FRE stage. For example, $L_{bb}^i$ denotes $i^{th}$ level of Backbone stage.   

\vspace{-0.1cm}
\subsection{The Backbone Stage}
\label{section:backbone}

The Backbone stage has $l$ levels denoted as $\{{ L}_{bb}^0, { L}_{bb}^1, \dots, { L}_{bb}^{l-1}\}$. At every level ${ L}_{bb}^i$, corresponding 3D points ${P}_{bb}^i$ (along with its features ${F}_{bb}^i$) are hierarchically down-sampled and passed to the next level ${ L}_{bb}^{i+1}$, i.e., $N_{bb}^i > N_{bb}^{i+1}$. We use the existing network PointNet++~\cite{pointnet2} as our backbone for its simpler design, multi-level nature of feature extraction and widespread use. However, other hierarchical feature extractors~\cite{densepoint, pointcnn, dgcnn} can also be used as backbone. Specifically, we use multi-Scale Grouping (MSG) version of PointNet++~\cite{pointnet2} to capture local neighborhood information at different scales. 

First, we briefly introduce our backbone (PointNet++~\cite{pointnet2}) and then mention the key modifications made by us for efficient processing. At every level ${ L}_{bb}^i$ of our backbone, first, $N_{bb}^{i+1}$ center points are sampled using farthest point sampling (FPS)~\cite{pointnet2} technique. Points within a sphere of radius $d$, centered on these points are selected and a shared local PointNet~\cite{pointnet} aggregates their features. Multiple radii are used to capture local shape variations and contextual cues. The mathematical formulation of grouping and feature aggregation can be written as:

\vspace{-0.5cm}
\begin{equation}
    f^{i+1}_{bb}(p) = \sigma(\{\phi(f^{i}_{bb}(q)); \forall \  q \in ||q-p||_2 \leq d\})
\label{eq:pointnet++}
\end{equation}
\vspace{-0.5cm}

where $p \in {P}_{bb}^{i+1}$ is the center point, $q \in {P}_{bb}^i$ is within the distance $d$ from $p$, $\phi (\cdot)$ is a shared point-wise feature transformation layer (shared-MLPs) and $\sigma (\cdot)$ is a symmetric aggregation function (a max pooling layer).

Next, we perform two modifications in the Backbone for reducing parameters as well as for efficient gradient propagation: 1) To reduce the parameters, we replace feature transformation layers (MLPs) with grouped convolutions~\cite{alexnet} inspired by~\cite{densepoint}, where the input channels are divided into $N_g$ groups and convolved separately. 2) To facilitate direct gradient propagation in the backbone network, we add residual connections~\cite{resnet} across the grouped convolution layers. 
\eqnmention \ref{eq:pointnet++} is modified to the utilize residual connections as follows:

\vspace{-0.7cm}
\begin{align}
    f^{i+1}_{bb}(p) = \sigma(\{\phi&(f_{bb}^{i}(q)) + r(f^{i}_{bb}(q));\nonumber \\&\forall \  q \in ||q-p||_2 \leq d\})
\label{eq:2}
\end{align}

Henceforth, $\phi(\cdot)$ denotes a point-wise feature transformation function with grouped convolutions and $r(\cdot)$ denotes a residual function unless otherwise specified. In the Backbone stage, $r(\cdot)$ is implemented as an identity function. 

In the final backbone level ($L^{l-1}_{bb}$), the features are aggregated using a global max pooling function and passed to the FCR stage.

\subsection{Feature Cross-Referencing (FCR) Stage}
\label{section:fcr}

The motivation of this stage is to preserve and refine the features from backbone by letting the low-level and high-level abstract features interact with each other. Such interactions between multi-level abstract features are proven crucial in segmentation~\cite{unet} as well as recognition~\cite{fishnet} tasks. 
To achieve this, we carefully design learning blocks to merge multi-level abstract features. 

\begin{figure}
\vspace{-0.3cm}
\begin{center}
    \includegraphics[width=\linewidth]{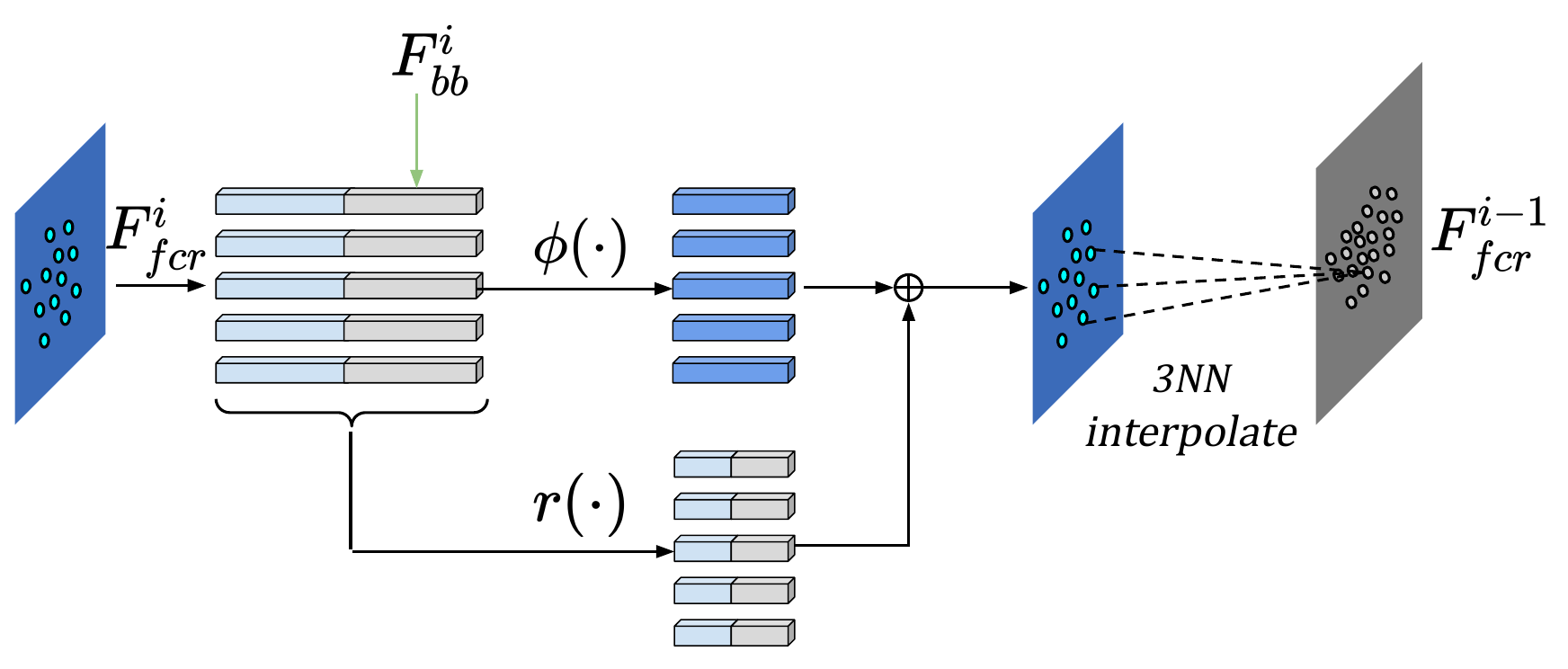}
\end{center}
\vspace{-0.5cm}
   \caption{The functionality of level $i$ in FCR stage. First, features from the same level of Backbone ($f_{bb}^i$) and FCR stage ($f_{fcr}^i$) are concatenated. Then, concatenated features are passed through a point-wise transformation ($\phi(\cdot)$) and reduction ($r(\cdot)$) layers. The output features from these two layers are added and further upsampled using 3NN interpolation~\cite{pointnet2}) technique to pass to next level.}
\label{fig:fcr}
\vspace{-0.4cm}
\end{figure}

Let $\{F_{bb}^0, F_{bb}^1, \dots, F_{bb}^{l-1}\}$ be the features from $\{{ L}_{bb}^0, { L}_{bb}^1, \dots, { L}_{bb}^{l-1}\}$ levels of Backbone stage. As Backbone stage has $l$ levels of hierarchy to extract multi-level abstract features, FCR stage also has $l$ levels to merge and refine features from the Backbone. This design is similar to a decoder network in~\cite{unet}.
FCR levels are numbered from $l-1$ to $0$ such that $N_{fcr}^{i} = N_{bb}^i$, $N_{fcr}^{i} < N_{fcr}^{i-1}$, i.e., the number of points of Backbone and FCR stages at a particular abstraction level are kept equal. The final level backbone features $F^{l-1}_{bb}$ are directly considered as $F^{l-1}_{fcr}$.
Next, the functionality of every other FCR level can be defined in three steps (\figmention \ref{fig:fcr}) as follows:

\vspace{-0.5cm}
\begin{align}
     F_{cat}^i &= concat(F_{fcr}^{i}, F_{bb}^{i}) \label{eq:fcr_cat} \\
     F_{ref}^i &= \phi (F_{cat}^i) + r(F_{cat}^i) \label{eq:fcr_reduce} \\
     F_{fcr}^{i-1} &= upsample(F_{ref}^i) \label{eq:fcr_upsample}
\end{align}

In the first step (\eqnmention \ref{eq:fcr_cat}), the point-wise feature vectors at the level $i$ from Backbone and FCR stages are concatenated together ($F_{cat}^i$). 
Next, in the second step (\eqnmention \ref{eq:fcr_reduce}), to get refined features ($F_{ref}^i$) from concatenated features, $F_{cat}^i$ is passed through a point-wise transformation function $\phi(\cdot)$ and a residual function $r(\cdot)$. Here, $r(\cdot)$ is implemented as a reduction function as explained below.

\vspace{-0.4cm}
\paragraph{Reduction function $r(\cdot)$:} 
It takes a feature vector $f$ of $D$ dimension as input and aggregates the features by summing up $k$ adjacent feature dimensions of the feature vector. Thus the resulting feature dimension is $\floor{\frac{D}{k}}$ ($D$ is chosen to be divisible by $k$). This reduction function serves two purposes: 1) to reduce the feature dimension without involving new parameters, and 2) to backpropagate unmodified gradients to earlier layers. Mathematically, the reduction function $r(\cdot)$ can be written as:

\vspace{-0.6cm}

\begin{equation}
    r(f) = \left[\sum_{j=1}^{k}f[j], \sum_{j=k+1}^{2k}f[j],...,\sum_{j=D-k}^{D}f[j]\right]
    \label{eq:reduction}
\end{equation}

Here, $f[j]$ denotes $j^{th}$ dimension of feature vector $f$.

The output features from two transformations ($\phi(\cdot), r(\cdot)$) are added channel-wise to form refined features $F_{ref}^i$. 
In the third step (\eqnmention \ref{eq:fcr_upsample}), the refined features are up-sampled to the high resolution points at next level $F_{fcr}^{i-1}$ by means of three nearest neighbor (3NN) interpolation method~\cite{pointnet2}.

\subsection{Feature Re-Encoding (FRE) Stage}
\label{section:fre}

\begin{figure}
\vspace{-0.4cm}
\begin{center}
    \includegraphics[width=\linewidth]{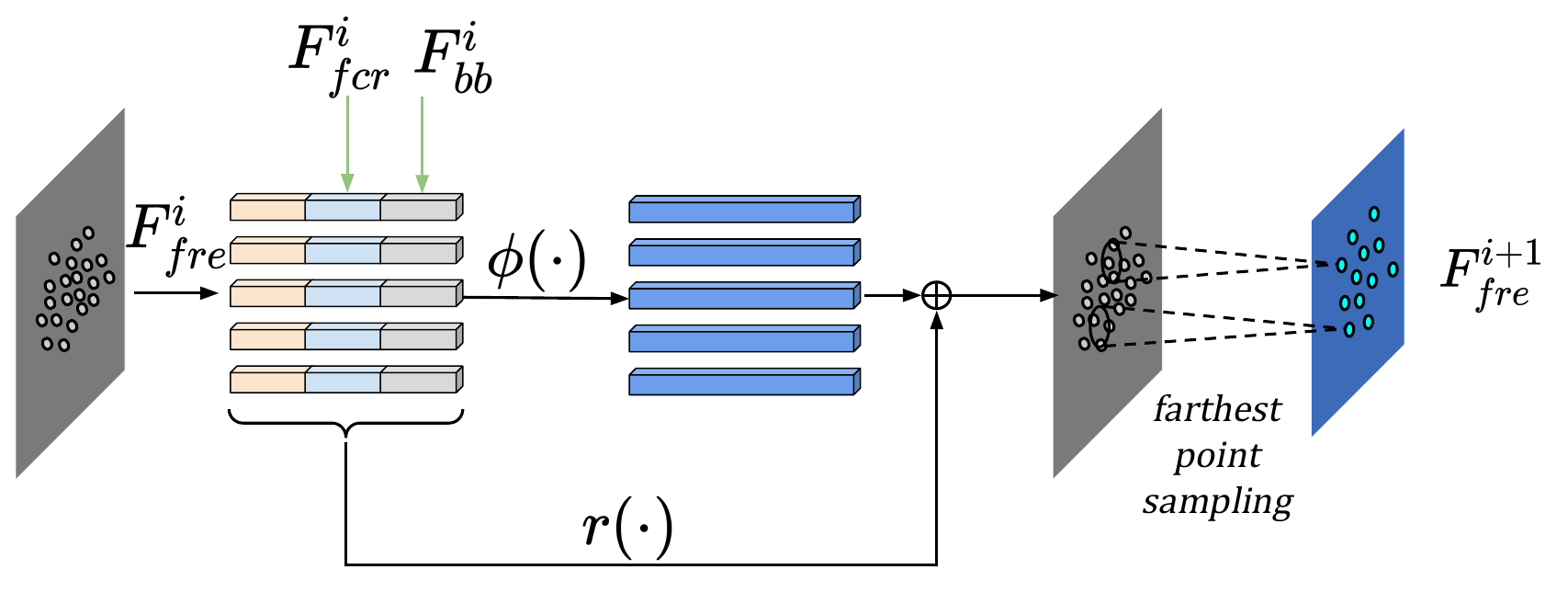}
\end{center}
\vspace{-0.5cm}
   \caption{The functionality of level $i$ in FRE stage. The features from all three stages (Backbone, FCR, FRE) are concatenated and passed through a residual point-wise transformation layer ($\phi(\cdot) + r(\cdot)$). $r(\cdot)$ is an identity function. The output features are propagated to the next level by `sampling, grouping, and feature averaging'.}
\label{fig:fre}
\vspace{-0.5cm}
\end{figure}

In the pipeline so far, the multi-level features from the Backbone stage have been cross-referenced and merged by FCR stage to capture local and global contextual cues. Similar to the Backbone and FCR stages, FRE stage has $l$ levels. In each of them, the multi-level features from both Backbone and FCR stages are combined and re-encoded to summarize multi-abstract features. The functionality of each FRE level is written in three steps (\figmention \ref{fig:fre}) as follows:

\vspace{-0.5cm}
\begin{align}
     F_{cat}^i &= concat(F_{fre}^{i}, F_{fcr}^{i}, F_{bb}^{i}) \label{eq:fre_cat} \\
     F_{ref}^i &= \phi (F_{cat}^i) + r(F_{cat}^i) \label{eq:fre_residual} \\
     F_{fre}^{i+1} &= downsample(F_{ref}^i) \label{eq:fre_downsample}
\end{align}

In the first step (\eqnmention \ref{eq:fre_cat}), features of the same abstraction level from all stages (Backbone, FCR, FRE) are concatenated together to get concatenated features $F_{cat}^i$. In the second step (\eqnmention \ref{eq:fre_residual}), concatenated features $F_{cat}^i$ are passed through the point-wise shared-MLP layers ($\phi(\cdot$)) with grouped convolutions along with the residual function ($r(\cdot)$) to obtain the refined features $F_{ref}^i$. Here, similar to the Backbone stage, $r(\cdot)$ is implemented as an identity function. Further, in the third step (\eqnmention \ref{eq:fre_downsample}), to pass $N_{fre}^{i+1}$ points to $({i+1})^{th}$ level (where $N_{fre}^{i+1} < N_{fre}^{i}$), we select $N_{fre}^{i+1}$ center points using the farthest point sampling technique and the 3D points are grouped in a Euclidean metric space around the selected centers. Then, the features corresponding to these points in each group are aggregated and passed to the next level. This process of grouping and aggregation is similar to the Backbone stage.

The final level features of FRE stage $F_{fre}^{l-1}$ are used for predicting the target tasks (shape classification in Sec. \ref{section:cls} and semantic segmentation in Sec. \ref{section:partnet}). The specification of exact architectures for target tasks is given in the supplementary material.

\subsection{Unique characteristics of MARNet}
\label{section:gradients}

The two key characteristics that make our network unique are:

\vspace{-0.5cm}
\paragraph{Multi-Abstraction Feature Aggregation:}
Our network aggregates the features of three granularity (from Backbone, FCR, FRE) in Re-encoding (FRE) stage. Hence, the final classification layer receives refined features from every level of abstraction. Further, this rich and diverse information can be used by the downstream classification/segmentation layers to precisely classify/segment the objects/parts.

\vspace{-0.5cm}
\paragraph{Unimpeded Gradients Propagation:} Exploding/Vanishing gradient~\cite{vanishinggradients} is a well-known problem in deep networks that makes the network training unstable. In our network, the corresponding abstraction levels in all three stages are connected to the final (classification/segmentation) layer through direct connections. For instance, Backbone features from level $L_{bb}^i$ is merged to FCR stage's level $L_{fcr}^i$ through feature concatenation (\eqnmention \ref{eq:fcr_cat} \& \ref{eq:fcr_reduce}) and reduction function, which does not involve any parameters. Similarly, FCR stage's features from level $L_{fcr}^i$ are passed via FRE stage through a concatenation and identity residual layer (\eqnmention \ref{eq:fre_cat} \& \ref{eq:fre_residual}) to the final layer.
In this way, every layer of our network receives the gradients directly from the loss function in an unmodified way.

\subsection{Implementation Details:}

MARNet is implemented using Pytorch~\cite{paszke2017automatic} framework. We optimize the models using Adam optimizer~\cite{adam} with following hyperparameters:  initial learning rate = 0.001, weight decay = 0.01, batch size = 32. The learning rate is decayed by multiplying with 0.7 after every 20 epochs. We train our model with an Nvidia GTX 1080Ti GPU. We elaborate on details about the models and training in the supplementary material.

\begin{table}
\small
\begin{center}
\begin{tabular}{|l|c|c|c|}
\hline

\textbf{Method} & \textbf{\#points} & \textbf{OA (\%)} & \textbf{mcA (\%)}     \\ \hline

PointNet~\cite{pointnet}	& 1k	& 89.2  & 86.2\\
SO-Net~\cite{sonet}	& 1k	& 89.4  & -\\
SCN~\cite{shapecontextnet}	& 1k	& 90.0  & 87.6\\
Kd-Net(depth=10)~\cite{kdnet} & 1k & 90.6 & 86.3 \\
PointNet++~\cite{pointnet2} & 1k   &  90.7  & - \\
Spec-GCN~\cite{specgcn}	& 1k	& 91.8  & -\\
DGCNN~\cite{dgcnn}	& 1k	& 92.2  & 90.2\\
PointCNN~\cite{pointcnn}	& 1k	& 92.2  & 88.1\\
PCNN~\cite{pcnn}	& 1k	& 92.3  & -\\
DensePoint~\cite{densepoint}	& 1k	& 93.2  & -\\
GeoCNN	~\cite{geocnn} & 1k	& 93.4   & {\bf 91.1} \\
{\bf Ours}	& {\bf 1k}	& {\bf 93.9}  & {\bf 91.1} \\ \hline

SO-Net~\cite{sonet}	& 2k	& 90.9  & 87.3\\
Kd-Net(depth=15)~\cite{kdnet}	& 32k	& 91.8  & 88.5\\
PointNet++~\cite{pointnet2}	& 5k	& 91.9  & -\\
Spec-GCN~\cite{specgcn}	& 2k	& 92.1  & -\\
SpiderCNN~\cite{spidercnn}	& 5k	& 92.4  & -\\
SO-Net~\cite{sonet}	& 5k	& 93.4  & 90.8\\\hline

\end{tabular}
\end{center}
\vspace{-0.5cm}
\caption{Classification Results on ModelNet40 dataset. Here, \textit{OA:} Overall Accuracy, \textit{mcA:} mean class Accuracy. (best results are in \textbf{bold})}
\label{tab:modelnet40}
\vspace{-0.2cm}
\end{table}

\section{Experiments}
\label{sec:experiments}
We validate the effectiveness of MARNet through rigorous experimentation on the tasks, namely: 1) Shape classification and 2) Coarse-to-fine grained semantic segmentation.

\subsection{Shape Classification}
\label{section:cls}

\begin{table}
\small
\begin{center}
\begin{tabular}{|l|c|c|c|}
\hline
\textbf{Method} & \textbf{\#points} & \textbf{OA (\%)} & \textbf{mcA} (\%)     \\ \hline

ECC~\cite{eccnet}	    & 1k	& 90.8 & 90.0\\
Kd-Net(depth=10)~\cite{kdnet}	& 1k	& 93.3 & 92.8\\
KCNet~\cite{kcnet}	& 1k	& 94.4  & - \\
PCNN~\cite{pcnn}	& 1k	& 94.9  & - \\
DensePoint~\cite{densepoint}	& 1k	& {\bf 96.6}  & -\\
{\bf Ours}        & {\bf 1k}    & 96.1  & {\bf 95.9}\\ \hline
Kd-Net(depth=15)~\cite{kdnet}	& 32k	& 94.0 & 93.5\\
SO-Net~\cite{sonet}	& 2k	& 94.1  & 93.9 \\
SO-Net~\cite{sonet}	& 5k	& 95.7  & 95.5 \\
\hline

\end{tabular}
\end{center}
\vspace{-0.4cm}
\caption{Classification results on ModelNet10 dataset. Here, \textit{OA:} Overall Accuracy, \textit{mcA:} mean class Accuracy. (best results are in \textbf{bold})}
\label{tab:modelnet10}
\vspace{-0.4cm}
\end{table}

\paragraph{Datasets and evaluation metrics:}
ModelNet40 and ModelNet10~\cite{modelnet40} are well known benchmarks for 3D point cloud recognition. Both datasets contain objects in the form of 3D CAD models. ModelNet40 contains 9843 train objects and 2468 test objects from 40 different object categories. ModelNet10 has 3991 train samples and 908 test samples divided into 10 object categories.
We evaluate the performance of our model with two evaluation metrics: 1) Overall Accuracy (\textit{OA}) - the ratio number of correctly predicted objects to the total number of objects in the dataset, 2) mean class accuracy (\textit{mcA}) - the average of the ratio of correctly predicted objects within a class to the total number of objects in the class. The \textit{OA} metric measures the overall performance of the model, whereas \textit{mcA} metric is more robust to the data imbalance within the classes.

Following~\cite{geocnn,pointnet2}, during training, we uniformly sample 1024 points along with point normals as input to the model. To make a fair comparison with the state-of-the-art, we use the same data augmentation techniques used by~\cite{densepoint,kdnet} to anisotropically scale the point clouds randomly in the range of
$[0.66, 1.5]$ and translate them in range $[-0.2, 0.2]$. During testing, we use voting evaluation as followed in~\cite{densepoint, pointnet, pointnet2} to average predictions from 10 runs.

\vspace{-0.6cm}
\paragraph{Comparison with state-of-the-art methods:}
The performance comparison with the state-of-the-art~\cite{pointnet,sonet,shapecontextnet,kdnet,pointnet2,specgcn,dgcnn,pointcnn,pcnn,densepoint,geocnn} methods in the ModelNet40 dataset is shown in Tab. \ref{tab:modelnet40}. For convenience, we categorize the methods based on the input number of points for the model. First, we notice that MARNet significantly outperforms its backbone method PointNet++~\cite{pointnet2}, with an improvement of about +3\% in overall accuracy. This improvement with a large margin validates our hypothesis that allowing multi-abstraction features to refine each other and preserving them from earlier layers is important for point cloud analysis. Further, our method also outperforms all the previous state-of-the-arts. Specifically, MARNet improves the performance over GeoCNN~\cite{geocnn} and DensePoint~\cite{densenet} by +0.5\% and +0.7\% respectively on the overall accuracy. In terms of mean class accuracy (\textit{mcA}), we perform on-par to GeoCNN~\cite{geocnn} and outperform all other methods in ModelNet40 dataset.
Experiments on ModelNet10 dataset (Tab. \ref{tab:modelnet10}) shows that MARNet performs on-par with DensePoint~\cite{densepoint} and outperform all other state-of-the-arts in both the metrics ({\it OA} and {\it mcA}).

\subsection{Coarse-to-fine grained Semantic Segmentation}
\label{section:partnet}

\begin{table*}[h!]
\begin{center}
\setlength{\tabcolsep}{3pt}
\resizebox{\textwidth}{!}
{\begin{tabular}{|l|l|llllllllllllllllllllllll|}
\hline
Method & Avg  & Bag  & Bed  & Bott & Bowl & Chair & Clock & Dish & Disp & Door & Ear  & Fauc & Hat  & Key  & Knife & Lamp & Lap  & Micro & Mug  & Ref & Scis & Stora & Table & Trash & Vase \\
\hline
P1     & 57.9 & 42.5 & 32   & 33.8 & 58   & 64.6  & 33.2  & 76   & 86.8 & 64.4 & 53.2 & 58.6 & 55.9 & 65.6 & 62.2  & 29.7 & 96.5 & 49.4  & 80   & 49.6 & 86.4 & 51.9  & 50.5  & 55.2  & 54.7 \\
P2     & 37.3 & -   & 20.1 & -    & -    & 38.2  & -     & 55.6 & -    & 38.3 & -    & -    & -    & -    & -     & 27   & -    & 41.7  & -    & 35.5 & -    & 44.6  & 34.3  & -     & -    \\
P3     & 35.6 & -    & 13.4 & 29.5 & -    & 27.8  & 28.4  & 48.9 & 76.5 & 30.4 & 33.4 & 47.6 & -    & -    & 32.9  & 18.9 & -    & 37.2  & -    & 33.5 & -    & 38    & 29    & 34.8  & 44.4 \\

\hline
P Avg    & 51.2 & 42.5 & 21.8 & 31.7 & 58   & 43.5  & 30.8  & 60.2 & 81.7 & 44.4 & 43.3 & 53.1 & 55.9 & 65.6 & 47.6  & 25.2 & 96.5 & 42.8  & 80   & 39.5 & 86.4 & 44.8  & 37.9  & 45    & 49.6 \\
\hline
P+1    & 65.5 & 59.7 & 51.8 & 53.2 & 67.3 & 68    & 48    & 80.6 & 89.7 & 59.3 & 68.5 & 64.7 & 62.4 & 62.2 & 64.9  & 39   & 96.6 & 55.7  & 83.9 & 51.8 & 87.4 & 58    & 69.5  & 64.3  & 64.4 \\
P+2    & 44.5 & -    & 38.8 & -    & -    & 43.6  & -     & 55.3 & -    & 49.3 & -    & -    & -    & -    & -     & 32.6 & -    & 48.2  & -    & 41.9 & -    & 49.6  & 41.1  & -     & -    \\
P+3    & 42.5 & -    & 30.3 & 41.4 & -    & 39.2  & 41.6  & 50.1 & 80.7 & 32.6 & 38.4 & 52.4 & -    & -    & 34.1  & 25.3 & -    & 48.5  & -    & 36.4 & -    & 40.5  & 33.9  & 46.7  & 49.8 \\
\hline
P+ Avg    & 58.1 & 59.7 & 40.3 & 47.3 & 67.3 & 50.3  & 44.8  & 62   & 85.2 & 47.1 & 53.5 & 58.6 & 62.4 & 62.2 & 49.5  & 32.3 & 96.6 & 50.8  & 83.9 & 43.4 & 87.4 & 49.4  & {\bf48.2}  & 55.5  & 57.1 \\
\hline
S1     & 60.4 & 57.2 & 55.5 & 54.5 & 70.6 & 67.4  & 33.3  & 70.4 & 90.6 & 52.6 & 46.2 & 59.8 & 63.9 & 64.9 & 37.6  & 30.2 & 97   & 49.2  & 83.6 & 50.4 & 75.6 & 61.9  & 50    & 62.9  & 63.8 \\
S2     & 41.7 & -    & 40.8 & -    & -    & 39.6  & -     & 59   & -    & 48.1 & -    & -    & -    & -    & -     & 24.9 & -    & 47.6  & -    & 34.8 & -    & 46    & 34.5  & -     & -    \\
S3     & 37   & -    & 36.2 & 32.2 & -    & 30    & 24.8  & 50   & 80.1 & 30.5 & 37.2 & 44.1 & -    & -    & 22.2  & 19.6 & -    & 43.9  & -    & 39.1 & -    & 44.6  & 20.1  & 42.4  & 32.4 \\
\hline
S Avg    & 53.6 & 57.2 & 44.2 & 43.4 & 70.6 & 45.7  & 29.1  & 59.8 & 85.4 & 43.7 & 41.7 & 52   & 63.9 & 64.9 & 29.9  & 24.9 & {\bf97}   & 46.9  & 83.6 & 41.4 & 75.6 & 50.8  & 34.9  & 52.7  & 48.1 \\
\hline
C1     & 64.3 & 66.5 & 55.8 & 49.7 & 61.7 & 69.6  & 42.7  & 82.4 & 92.2 & 63.3 & 64.1 & 68.7 & 72.3 & 70.6 & 62.6  & 21.3 & 97   & 58.7  & 86.5 & 55.2 & 92.4 & 61.4  & 17.3  & 66.8  & 63.4 \\
C2     & {\bf46.5} & -    & 42.6 & -    & -    & 47.4  & -     & 65.1 & -    & 49.4 & -    & -    & -    & -    & -     & 22.9 & -    & 62.2  & -    & 42.6 & -    & 57.2  & 29.1  & -     & -    \\
C3     & {\bf46.4} & -    & 41.9 & 41.8 & -    & 43.9  & 36.3  & 58.7 & 82.5 & 37.8 & 48.9 & 60.5 & -    & -    & 34.1  & 20.1 & -    & 58.2  & -    & 42.9 & -    & 49.4  & 21.3  & 53.1  & 58.9 \\
\hline
C Avg    & 59.8 & 66.5 & {\bf46.8} & 45.8 & 61.7 & {\bf53.6}  & 39.5  & {\bf68.7} & 87.4 & {\bf50.2} & 56.5 & {\bf64.6} & {\bf72.3} & 70.6 & 48.4  & 21.4 & {\bf97}   & {\bf59.7}  & {\bf86.5} & {\bf46.9} & {\bf92.4} & {\bf56}    & 22.6  & {\bf60}    & {\bf61.2} \\
\hline
Ours1	&{\bf67.3}  &88.3	&56.7	&42.4	&74	    &45.8	&71.3	&80.6	&91.8	&60.1	&67.3	&65.8	&67.4	&70.7	&60.7	&49.3	&97	    &49.9	&86.1	&49.7	&90.2	&55.8	&72.4	&58.9	&63.6 \\

Ours2	&45.4  &-	 &35.4 &-	&-			&48 &-	   &63.8	&-	&42	&-	&-	&-	&-	&-	&40	&-	&49.5	&-	&42.6	&-	&45	 &42	&-	&- \\

Ours3	&43.6	&- &30.3	&58.5	&-	&41.3	&16	&50.7	&83.1	&45.4	&46.5	&50.8	&-	&-	&48.5	&27.2	&-	&64	&-	&38.4	&-	&40.2	&16	&37.7	&46.1	\\
\hline
Ours Avg &{\bf60.6} &{\bf88.3}	&40.8	&{\bf50.5}	&{\bf74}	&43.5	&{\bf45.1}	&65	&{\bf87.5}	&49.2	&{\bf56.9}	&58.3	&67.4	&{\bf70.7}	&{\bf54.6}	&{\bf38.8}	&{\bf97}	&54.5	&86.1	&43.6	&90.2	&47	&43.5	&48.3	&54.8	\\
\hline
\end{tabular}}
\end{center}
\vspace{-0.5cm}
\caption{Coarse-to-fine grained Semantic Segmentation results on PartNet dataset (part-category mIoU \%). The five methods are P:PointNet~\cite{pointnet}, P+:PointNet++~\cite{pointnet2}, S:SpiderCNN~\cite{spidercnn}, C:PointCNN~\cite{pointcnn}  and ours:MARNet. These methods are compared on three levels of segmentation 1:coarse-, 2:middle- and 3:fine-grained for 24 different categories. The average across the three levels and the average across the shape categories reveal that MARNet performs better or on-par with other methods. (best values are highlighted).}
\label{tab:partseg_results}
\vspace{-0.1cm}
\end{table*}

\begin{figure*}
\vspace{-0.3cm}
\begin{center}
    \includegraphics[width=\linewidth]{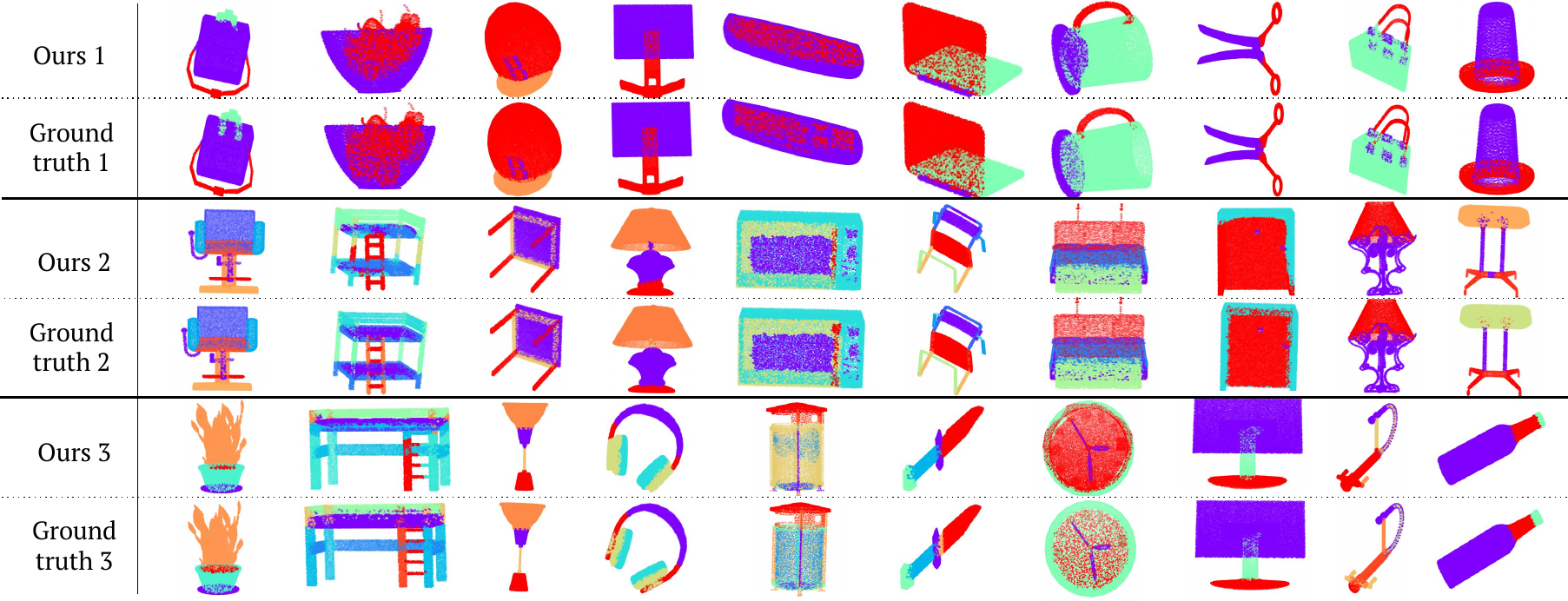}
\end{center}
\vspace{-0.6cm}
   \caption{Qualitative results on PartNet Coarse-to-fine grained semantic segmentation. Top, Middle, and Bottom rows show MARNet prediction vs ground truths of 1:coarse-, 2:middle-, and 3:fine-grained segmentation respectively.}
\label{fig:qualitative}
\vspace{-0.5cm}
\end{figure*}

\paragraph{Dataset and evaluation metric: }
PartNet~\cite{PartNet} is a semantic (part) segmentation dataset consisting of three levels of shapes namely: 1) coarse, 2) middle, and 3) fine-grained. It is build on top of ShapeNet~\cite{ShapeNet} and contains 26,671 3D models from 24 different object categories with 573,585 annotated part instances.

We evaluate our model on all three levels of segmentation. Following~\cite{PartNet}, we use part-category mIOU (in \%) as the evaluation metric and compare our results with the published state-of-the-art methods: PointNet~\cite{pointnet}, PointNet++~\cite{pointnet2}, SpiderCNN~\cite{spidercnn}, PointCNN~\cite{pointcnn}. In our experiments, we train individual networks for segmentation of different object categories and levels (coarse, middle and fine-grained), as followed in~\cite{PartNet}. Precisely, we train 50 separate networks: 24 for coarse-level, 9 for middle-level and 17 for fine-grained level.

\vspace{-0.4cm}
\paragraph{Comparison with state-of-the-art methods:}
\tabmention \ref{tab:partseg_results} compares the performance of MARNet with other state-of-the-art methods on three different tasks: coarse-, middle- and fine-grained semantic segmentation of objects. The results are averaged across the three tasks as well as across the categories for each of the tasks. MARNet performs significantly better on the coarse-grained task with +1.8\% improvement over the previous best performing method, PointNet++~\cite{pointnet2}. On average, across all three tasks, our model performs better than the previous state-of-the-arts with +2.5\% and +0.8\% over PointNet++~\cite{pointnet2} and PointCNN~\cite{pointcnn}, respectively. The qualitative results shown in \figmention \ref{fig:qualitative} illustrates the effectiveness of MARNet in semantic segmentation task.

\vspace{-0.2cm}
\section{Ablation Study}
\label{sec:ablation}

We conduct several ablation studies to verify the architectural choices adopted for MARNet, its robustness to the variations in input data and its computation efficiency. Specifically, we conduct experiments to ablate on 1) various components in the network, 2) the number of input points to the network, and 3) noi2y input data. Further, we outline the advantages of MARNet in terms of memory and computational complexity (measured by the number of parameters and FLOPs, respectively).
All the ablation studies are conducted on ModelNet40~\cite{modelnet40} classification dataset.

\vspace{-0.2cm}
\subsection{Componentwise contributions in MARNet}
\vspace{-0.2cm}
In this section, we conduct experiments to dissect the contributions of different components in MARNet and show the results in \tabmention \ref{tab:ablate_components}. First, we train only the backbone network (modified PointNet++, as in \secmention \ref{section:backbone}) without data augmentation (DA) and observe the performance as 89.1\%. Adding data-augmentation (\secmention \ref{section:cls}) along with the Backbone model (Backbone+DA) improves the performance by +1.1\%. Next, we add FCR, FRE stages to formulate MARNet, but without residual function (MARNet(w/o $r(\cdot)$) model). This simple architecture add-on significantly improves the classification accuracy by +2.4\%. Such performance increase asserts that allowing multi-level features to interact with cross-referencing and re-encoding layers provides more contextual information. Next, we introduce residual functions $r(\cdot)$ in all stages of MARNet (as explained in \secmention \ref{section:backbone}, \ref{section:fcr}, \ref{section:fre}) to improve the training stability and avoid vanishing/exploding gradients. MARNet(w/ $r(\cdot)$) variant passes unmodified gradients from final classification layer to all levels in 3 stages and improves the performance further by +0.8\%. Incorporating a voting mechanism for testing (as followed in~\cite{densepoint,pointnet}) adds +0.5\% improvement to MARNet.

\begin{table}
\small
\vspace{-0.3cm}
\setlength{\tabcolsep}{3pt}
\begin{center}
\begin{tabular}{|l|c|c|c|c|c|}
\hline
\textbf{Model} & \textbf{DA} & \textbf{\makecell{point-wise\\trans.}} & \textbf{\makecell{residual\\trans.}} & \textbf{voting} & \textbf{OA(\%)}     \\
  & & $\phi(\cdot)$ & $r(\cdot)$ & & \\
\hline
Backbone-only & & & & & 89.1 \\
Backbone+DA & \checkmark & & & & 90.2\\
MARNet(w/o $r(\cdot)$) & \checkmark &\checkmark & & & 92.6 \\
MARNet(w/ $r(\cdot)$) &\checkmark &\checkmark &\checkmark & & 93.4 \\
MARNet &\checkmark &\checkmark &\checkmark &\checkmark & {\bf 93.9} \\ \hline
\end{tabular}
\end{center}
\vspace{-0.5cm}
\caption{Ablation studies on various components of MARNet. Backbone-only = modified PointNet++, Backbone+DA = Baseline (modified PointNet++ backbone as in \secmention \ref{section:backbone}) with Data Augmentation (DA), MARNet(w/o $r(\cdot)$) = MARNet without residual connections in all three stages, MARNet(w/ $r(\cdot)$) = including the residual connections and MARNet = complete MARNet evaluated with voting evaluation, \textit{OA} = Overall Accuracy}.
\label{tab:ablate_components}
\vspace{-0.4cm}
\end{table}

\begin{figure}[t]
\vspace{-0.1cm}
\begin{subfigure}{0.48\linewidth}
\begin{center}
\includegraphics[trim= 0 10 0 0, clip,width=\linewidth]{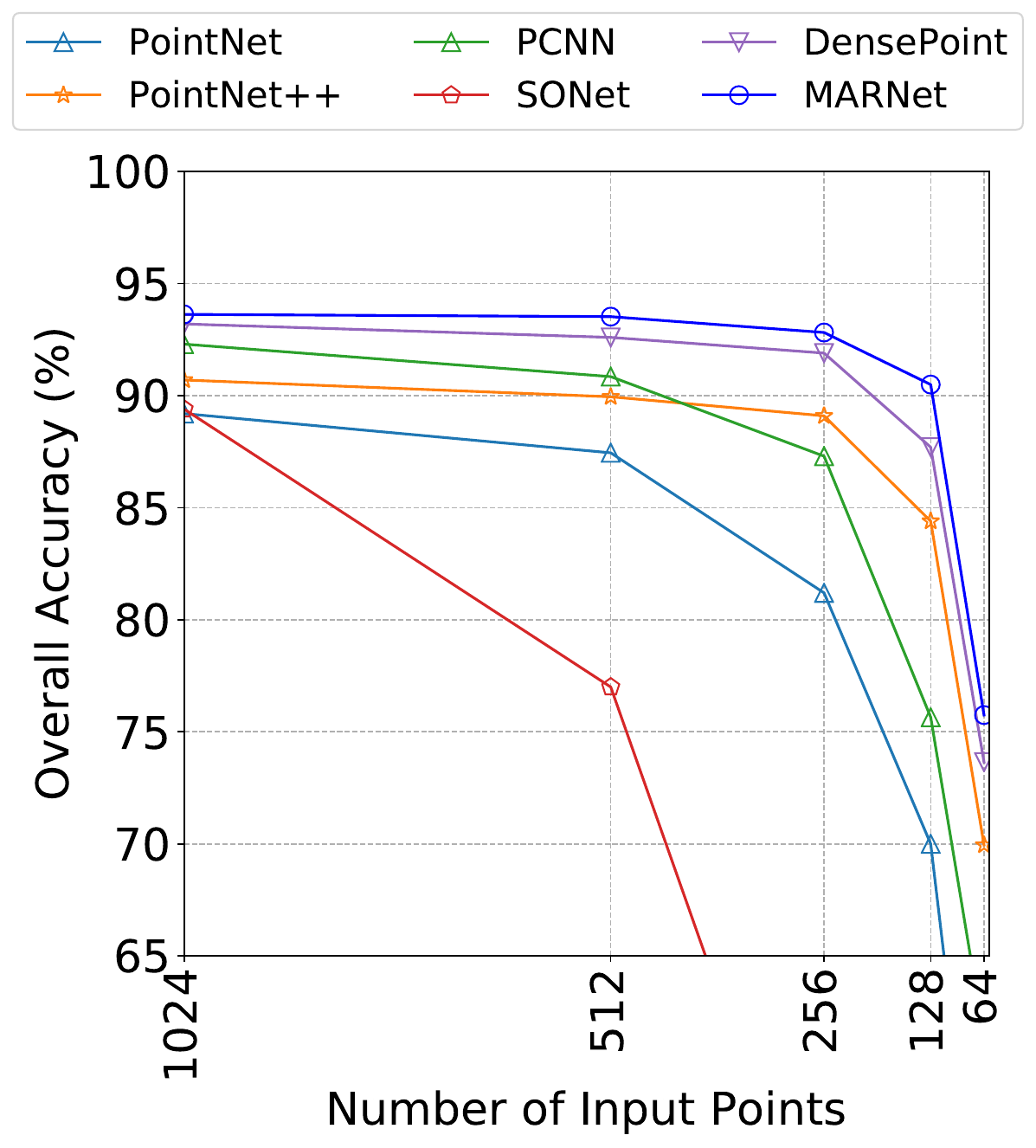}
\vspace{-0.5cm}
\caption{}
\label{fig:ablationgraph1}
\end{center}
\end{subfigure}\hfill
\begin{subfigure}{0.5\linewidth}
\begin{center}
\includegraphics[width=\linewidth]{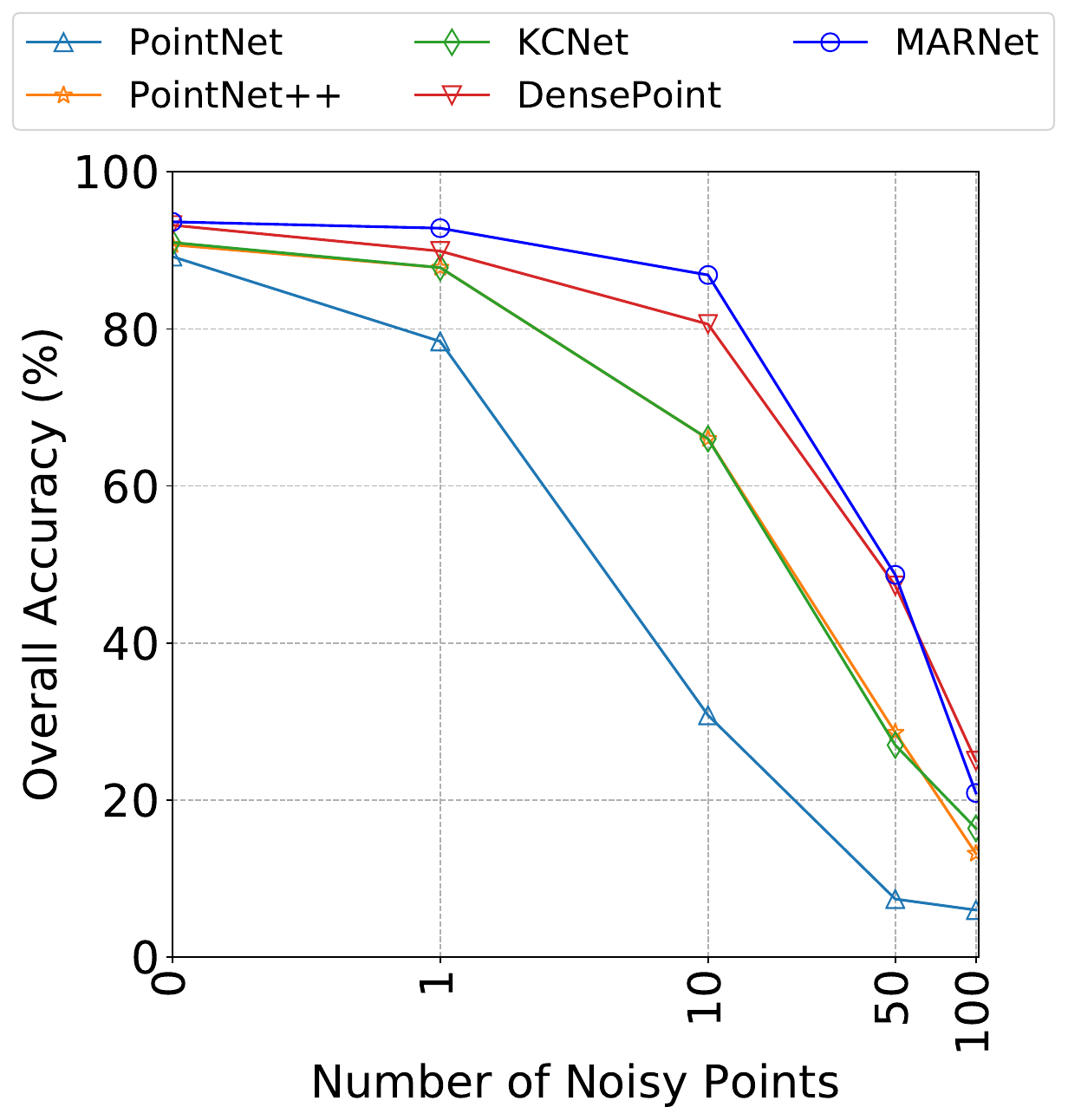}
\vspace{-0.5cm}
\caption{}
\label{fig:ablationgraph2}
\end{center}
\end{subfigure}
 \vspace{-0.4cm}
\caption{Performance comparison of various models on different number of input points (Left) and Noisy input points (Right). MARNet exhibits only minor degradation in performance compared to other state-of-the-art models.}
\vspace{-0.5cm}
\end{figure}

\subsection{Ablation on number of input points to the model}

The resolution of point clouds tends to affect the performance of the models. Typically, high-resolution point clouds may provide richer and unambiguous features to aid classification. To study the effect of the different number of input points, we test MARNet with varying number of input points selected by the farthest point sampling method. Specifically, we test MARNet on 512, 256, 128 and 64 input points and show the results in \figmention \ref{fig:ablationgraph1}. Compared to several state-of-the-art methods~\cite{pointnet,pointnet2,pcnn,sonet,dgcnn,densepoint}, MARNet exhibits robust performance with better classification accuracy than other methods even for sparse input points. For example, even with 128 points, MARNet's accuracy drops only by 2.9 \% from the overall accuracy. We hypothesize that the ability to preserve the features at all levels of abstraction helps MARNet achieve such robustness.

\subsection{Ablation on noisy input points}

3D point clouds acquired by real sensors could possess noisy points, unlike computer-generated CAD models. To analyze the robustness towards the noise, we simulate noisy points by including points sampled from a random uniform distribution between -1 and 1. We test MARNet and other state-of-the-art methods by including  1, 10, 50, and 100 noisy points to the actual 1024 input points. \figmention \ref{fig:ablationgraph2} reveals that MARNet performs on-par to DensePoint and better than other state-of-the-art methods.

\subsection{Model Complexity}
\tabmention \ref{tab:complexity} compares the complexities of different models in terms of the number of parameters and floating-point operations (FLOPs). MARNet achieves 93.9\% accuracy and has a competitive model complexity with 1.13M parameters and $\sim$1 GFLOPs. A significant number of parameters are contributed from our Backbone's (PointNet++~\cite{pointnet2}) MSG module. To avoid this, we propose a variant of MARNet, namely ``MARNet Lite", by replacing the MSG layers with a single scale grouping layer. The resulting MARNet Lite has significantly less number of FLOPs and parameters with only a negligible drop in the accuracy.

\begin{table}
\small
\begin{center}
\setlength{\tabcolsep}{3pt}
\begin{tabular}{|l|c|c|c|c|c|}
\hline
{\bf Model} & {\bf\#params} & {\bf \#FLOPs} & {\bf OA} \\ \hline 
PointNet~\cite{pointnet} &3.50M & 440M & 89.2 \\
KCNet~\cite{kcnet} &0.9M & - &91.0 \\
PointNet++~\cite{pointnet2} &1.48M & 1684M & 91.9 \\
SpecGCN~\cite{specgcn} &2.05M & 1112M & 92.1 \\
DGCNN~\cite{dgcnn} &1.84M & 2767M &92.2 \\
PointCNN~\cite{pointcnn} & {\bf 0.60M} & 1581M &92.2 \\
PCNN~\cite{pcnn} &8.20M & 294M &92.3 \\
DensePoint~\cite{densepoint} & 0.67M & 651M &93.2 \\ \hline
MARNet & 1.13M & 1040M &{\bf 93.9}  \\
MARNet Lite & 0.68M & {\bf 260M} &93.2 \\ \hline
\end{tabular}
\end{center}
\vspace{-0.5cm}
\caption{Comparison of different models in terms of the number of parameters and computational complexity (FLOPs). Here, \textit{OA} = Overall Accuracy}
\label{tab:complexity}
\vspace{-0.6cm}
\end{table}

\section{Conclusion}
\label{sec:conclusion}
\vspace{-0.2cm}
In this work, we proposed a novel three-stage deep learning architecture named MARNet to extract and refine the multi-level abstract features for point-cloud analysis. Not only MARNet aggregates features of different granularity, but it also preserves earlier layer features till the final layer and propagates unmodified gradients to earlier layers. With these unique advantages, MARNet shows promising improvements over state-of-the-arts in the tasks of 3D shape classification and semantic segmentation. As the design of MARNet architecture is generic, we intend to study the applicability of this in other 3D point cloud tasks such as point cloud registration and object detection in our future work.

{\small
\bibliographystyle{ieee_fullname}
\bibliography{references}
}

\newpage

\begin{center}
    \textbf{\Large Supplementary Material}\\
\end{center}

\setcounter{equation}{0}
\setcounter{figure}{0}
\setcounter{table}{0}
\setcounter{section}{0}

\renewcommand{\sectionautorefname}{\S}
\renewcommand{\subsectionautorefname}{\S}
\renewcommand{\subsubsectionautorefname}{\S}

\section{Outline}
In the supplementary section, we report additional ablation studies of our model on the following aspects:

\begin{itemize}

    \item varying number of groups ($N_g$) in Grouped convolution based MLP $\phi(\cdot)$ (\secmention \ref{sec:ablation_ngroup})
    \item varying number of levels in the backbone, Feature Cross-Referencing (FCR) and the Feature Re-Encoding (FRE) stages (\secmention \ref{sec:ablation_levels})
    \item Inference time and memory requirements (\secmention \ref{sec:computations})
\end{itemize} 

\begin{figure*}[h!]
\vspace{-0.3cm}
\begin{center}
    \includegraphics[width=\linewidth]{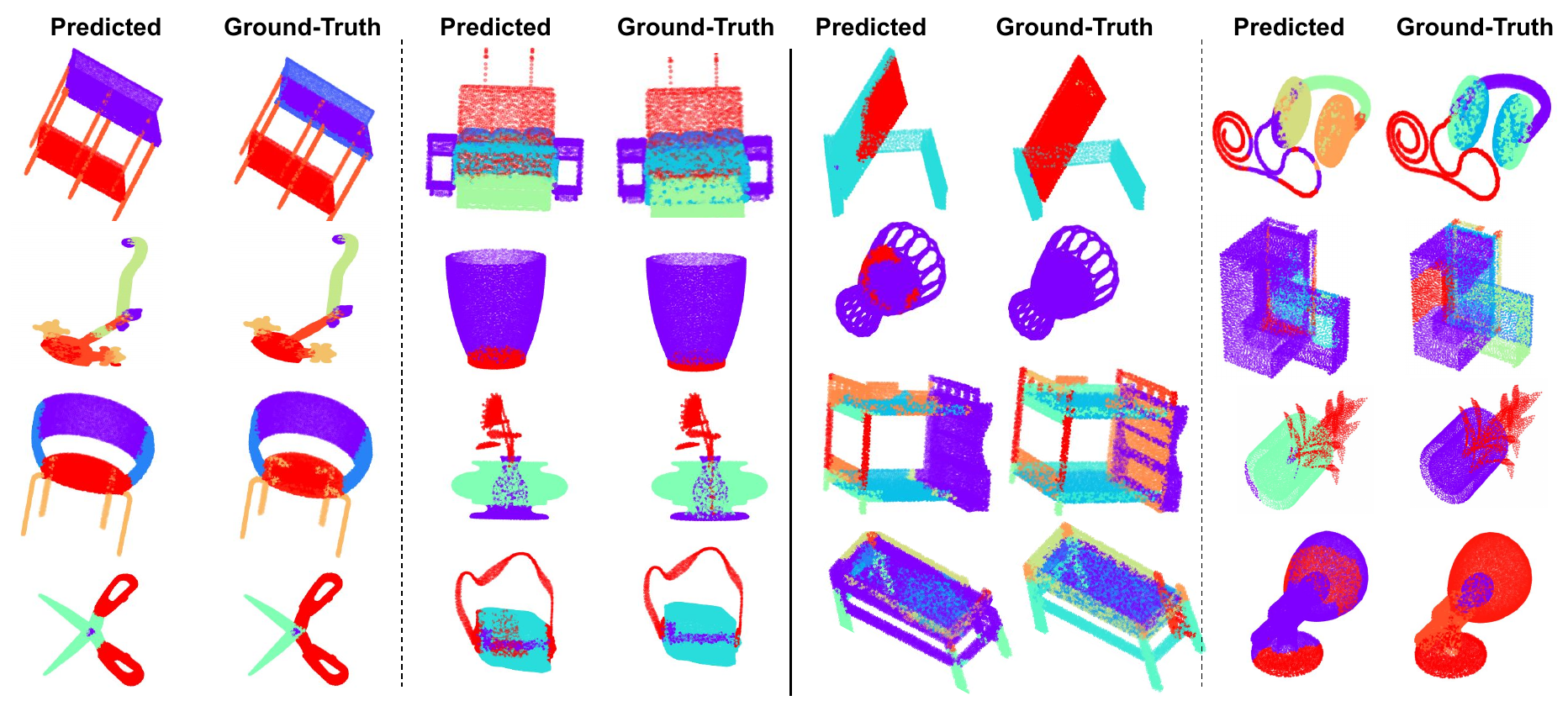}
\end{center}
\vspace{-0.6cm}
\caption{Additional qualitative results of MARNet on PartNet dataset. Columns 1 and 2 show the accurate predictions of MARNet with respect to the ground-truth. Columns 3 and 4 show some of the incorrect predictions.}
\label{fig:goodbad}
\end{figure*}

More qualitative results on PartNet part-segmentation can be found in \figmention \ref{fig:goodbad}. Additionally, we describe the exact network architecture of MARNet and its training details in \secmention \ref{sec:archdetails}, which will be of help to reproduce the work. 

\section{Ablation studies (contd.,)}
\subsection{Varying number of groups ($\boldsymbol{N_g}$) in Grouped Convolution based MLPs $\phi(\cdot)$}
\label{sec:ablation_ngroup}
In grouped convolution~\cite{alexnet}, input channels are divided into $N_g$ groups and convolution kernel is applied separately on them. Then, the output features are generated independently and concatenated to obtain the final output. This type of grouping imposes sparsity in the connections between the nodes and helps to reduce the model complexity in terms of number of parameters. 

\tabmention \hspace{-0.05cm}\ref{tab:ngroups} shows the influence of different $N_g$ on model complexity (\#parameters, \#floating point operations (FLOPs)) and accuracy of our model. We experiment with values of $N_g$ between 1 and 8. As $N_g$ increases, \#parameters and \#FLOPs decreases due to the sparsification of the connections between the nodes. We obtained the highest accuracy for $N_g = 2$ and thus, we select this configuration as our optimal model.

\begin{table}[h!]
\footnotesize
\begin{center}
\begin{tabular}{cccc}
\hline
$\boldsymbol{N_g}$	&\textbf{\#parameters}	&\textbf{\#FLOPs}	&\textbf{O.A. (\%)} \\ \hline
1 & 1.85M & 2080M & 93.2\\
2 & 1.13M & 1040M & {\bf 93.9}\\
4 & 0.84M & 580M  & 93.4\\
8 & 0.67M & 380M & 92.5\\ \hline
	
\end{tabular}
\end{center}
\vspace{-0.4cm}
\caption{MARNet complexity and performance for $N_g$ groups in grouped convolution. Here, \textit{OA} = Overall Accuracy.}
\label{tab:ngroups}
\vspace{-0.4cm}
\end{table}

\begin{table}
\footnotesize
\begin{center}
\begin{tabular}{lccc}
\hline
\textbf{\#levels}	&\textbf{\#parameters}	&\textbf{\#FLOPs}	&\textbf{O.A. (\%)} \\ \hline
BB = 3, FCR = 2, FRE = 2 &0.80M &1320M & 93.1\\
BB = 4, FCR = 3, FRE = 3 &1.13M &1040M & {\bf 93.9}\\
BB = 5, FCR = 4, FRE = 4 &2.06M &3060M & 93.5\\
BB = 6, FCR = 5, FRE = 5 &3.42M &8020M & 93.0\\ \hline
	
\end{tabular}
\end{center}
\vspace{-0.4cm}
\caption{MARNet complexity and performance for varying number of \#levels in MARNet. Here, \textit{BB, FCR, GRE} denote number of Backbone-, FCR- and FRE-levels respectively. The number of FC layers $=3$ in all cases. \textit{OA} = Overall Accuracy.}
\label{tab:nlevels}
\vspace{-0.4cm}
\end{table}

\begin{table}
\begin{center}
\begin{tabular}{cccc}
\hline
\textbf{model}	&\textbf{time}	&\textbf{memory}	&\textbf{O.A. (\%)} \\ \hline
MARNet & 24ms & 1951MB & 93.9\\
MARNet Lite & 8ms &  1055MB & 93.2 \\ \hline
	
\end{tabular}
\end{center}
\vspace{-0.4cm}
\caption{Inference time and GPU memory requirements for MARNet and MARNet Lite. Batch size of 32 is used. ``time'' denotes average inference time per test sample.}
\label{tab:inferencetime}
\vspace{-0.4cm}
\end{table}

\subsection{Varying number of levels in MARNet}
\label{sec:ablation_levels}
We report the model complexity (\#parameters, \#FLOPs) and its performance for different number of levels in MARNet. As mentioned in \secmention 3 of the main paper, MARNet has \textit{l} levels of abstraction (down-sampling) in the Backbone stage given by $\{{L}_{bb}^0, {L}_{bb}^1, \dots, { L}_{bb}^{l-1}\}$. For every level in the Backbone (except ${L}_{bb}^0$), one corresponding level is added in FCR and FRE stages respectively.

\tabmention \ref{tab:nlevels} gives the performance of MARNet by varying number of levels. MARNet with 10 levels (Backbone = 3, FCR = 2, FRE = 2, FC = 3) lacks sufficient capacity to learn multi-abstraction information and therefore, it under-performs. MARNet with 13 levels (Backbone = 4, FCR = 3, FRE = 3, FC layers = 3) gives the best accuracy of $93.9\%$. Further increasing the levels increases the model complexity and overfits to the training data while giving sub-par performance.

\subsection{Inference time and memory requirements}
\label{sec:computations}
We compare the computational requirements of our optimal models (MARNet and MARNet Lite with 13 levels: Backbone = 4, FCR = 3, FRE = 3, FC layers = 3) in \tabmention \ref{tab:inferencetime}. The test is executed on a Nvidia GTX 1080Ti GPU with batch size of 32.

We observe that MARNet is competitive in terms of inference time and memory requirement. However, MARNet Lite is 3 times faster than MARNet and achieves a near real-time inference speed with a little drop in accuracy.

\section{Network Architecture details}
\label{sec:archdetails}
In this section, we provide the exact settings used for training MARNet for different tasks to facilitate reproducibility. In particular, we give layer-by-layer configurations for the following
:
\begin{enumerate}
    \item MARNet for classification task (\tabmention \ref{tab:MARNetcls})
    \item MARNet for part-segmentation task (\tabmention \ref{tab:MARNetpartseg}) 
    \item MARNet lite for classification task (\tabmention \ref{tab:MARNetlite})
\end{enumerate}

Firstly, we explain the layer types used in our network. A brief description of the network components along with the hyper-parameters are explained below.

\paragraph{PointNetSetAbstraction:}
~\newline
\textit{\textbf{hyper-params:} (\#input channels, grouping radius, \#samples, [MLP output dims])}\\
We use Ball Point Query to select the points within the grouping radius and choose \#samples points among them to group and aggregate locally using PointNet~\cite{pointnet}. PointNet is implemented as series of MLPs with output dimensions given in \textit{MLP output dims} list. Max pooling is used to aggregate the local point features.

\paragraph{PointNetSetAbstractionMsg:}
~\newline
\textit{\textbf{hyper-params:} (\#input channels, [list of grouping radii], [list of \#samples], [[list of MLP output dims]])}\\
This layer is Multi-Scale Grouping (MSG) version of the \textit{PointNetSetAbstraction} layer. The grouping, transformation and aggregation operations of PointNet are applied for multiple radii in parallel and their features are concatenated in the end.

\paragraph{FeatureCrossReferencing:}
~\newline
\textit{\textbf{hyper-params:} (\#input channels, [MLP output dims])}\\
The backbone features are concatenated with the input of this layer. The features are passed through a set of MLPs and a reduction function before upsampling the points. To project the point features on the upsampled points using 3-nearest neighbor (3NN) interpolation technique. The 3NN technique interpolates the points according to the inverse of the weighted distance between the points. 

\paragraph{FeatureReEncoding:}
~\newline
\textit{\textbf{hyper-params:} (\#input channels, grouping radius, \#samples, [MLP output dims])}\\
The features from backbone and FCR layers are concatenated to this layer's input. Same set of sampled points from the backbone are passed to this layer. The grouping and aggregation operation is similar to the \textit{PointNetSetAbstraction} layer.

\paragraph{PointNetFeaturePropagation:}~\newline
\textit{\textbf{hyper-params:} (\#input channels, [MLP output dims])}\\
This layer is only employed in our part-segmentation network. It is the feature propagation layer of PointNet++~\cite{pointnet2}. The features from the encoder at the same level are concatenated, transformed and upsampled. The transformation is implemented as a series of MLPs.

\paragraph{FullyConnectedLayer:}~\newline
\textit{\textbf{hyper-params:} (\#input channels, \#output channels, dropout ratio \%)}\\
The FC layer conprises of a linear layer followed by a batch-normalization layer and a ReLU non-linearity. Dropout is applied after every layer except for the final layer which is used to make the predictions.

All the point-wise transformation layers in the Backbone, FCR and FRE stages use grouped convolutions with number of groups $N_g = 2$ throughout. 

In the backbone stage, apart from the input channels from the previous layer, 6 more channels (xyz+normals) are concatenated to features before applying PointNet~\cite{pointnet} locally.

\subsection{MARNet Classification Network}
Our ModelNet40 classification network consists of 13 levels in total: \#Backbone levels = 4, \#FCR levels = 3, \#FRE levels = 3, number of fully connected (FC) layers = 3. The details of our network design are mentioned in \tabmention \ref{tab:MARNetcls}.

\subsection{MARNet Part-Segmentation Network}
\tabmention \ref{tab:MARNetpartseg} outlines our network design for the PartNet part-segmentation task. It consists of 16 levels: \#Backbone levels = 4, \#FCR levels = 3, \#FRE levels = 3, PointNetFeaturePropagation layers = 4, and \#FC layers = 2.

\subsection{MARNet Lite for classification task}
This is the lite version of MARNet in terms of \#parameters and \#FLOPS. It comprises of the same layout as MARNet classification network except that the MSG layers in the backbone are replaced by the SSG (single scale grouping) layers. MARNet Lite's configuration is listed in \tabmention \ref{tab:MARNetlite}.

\begin{table*}[!htb]
\footnotesize
\begin{center}
{\begin{tabular}{ccccc}
\hline
\textbf{Layer}	&\textbf{Layer Type}	&\textbf{Layer Parameters}	&\textbf{Output}	&\textbf{S.C.} \\ \hline \hline

 & & \textbf{Backbone Stage} & & \\ \hline
BB1 &	PointNetSetAbstractionMsg &	$0, [0.1,0.2,0.4], [16,32,128], [[16, 16, 16],[16, 16, 16],[32, 32, 32]]$ &	$(64, 512)$ & \\	
BB2	& PointNetSetAbstractionMsg &	$64, [0.2,0.4,0.6], [32,64,128], [[32, 32, 32],[32, 32, 32],[64, 64, 64]]$	& $(128, 128)$ &	\\
BB3	& PointNetSetAbstractionMsg &	$128, [0.6,0.8,0.9], [64,96,128], [[64, 64, 64],[64, 64, 64],[128, 128, 128]]$	& $(256, 32)$ & \\	
BB4	& PointNetSetAbstraction	& $256, None, None, [256]$ & 	$(256, 1)$	& \\ \hline

 & & \textbf{Feature Cross Referencing Stage} & & \\ \hline
FCR1	&FeatureCrossReferencing	&$256, [128,128]$	    &$(128, 32)$	&    \\
FCR2	&FeatureCrossReferencing	&$128+256, [64, 64]$	&$(64, 128)$	&BB3 \\
FCR3	&FeatureCrossReferencing	&$64+128, [32, 32]$	    &$(32, 512)$	&BB2 \\ \hline
						
 & & \textbf{Feature Re-Encoding Stage} & & \\ \hline
FRE1	&FeatureReEncoding	&$32+64, 0.4, 32, [96, 96]$	        &$(96, 128)$	&BB1 \\
FRE2	&FeatureReEncoding	&$96+64+128, 0.8, 32, [288, 288]$	&$(288, 32)$	&BB2, FCR2 \\
FRE3	&FeatureReEncoding	&$288+128+256, None, None, [672,672]$	&$(672, 1)$	&BB3, FCR1 \\ \hline
						
 & & \textbf{Classification Stage} & & \\ \hline
FC1	&FullyConnectedLayer	&$672, 512, 0.4$	&$(512, 1)$ &	\\
FC2	&FullyConnectedLayer	&$512, 256, 0.5$	&$(256, 1)$ &	\\
FC3	&FullyConnectedLayer	&$256, \#classes$	&$(\#classes, 1)$ &	\\
\hline					
						
\end{tabular}}
\end{center}
\vspace{-0.4cm}
\caption{Network Architecture of MARNet Classification Network. Here, \textit{Output} = (\#output channels, \#output points), \textit{S.C.} = shortcut connections indicating features passed from previous layers. }
\label{tab:MARNetcls}
\vspace{-0.4cm}
\end{table*}

\begin{table*}[!htb]
\footnotesize
\begin{center}
{\begin{tabular}{ccccc}
\hline
\textbf{Layer}	&\textbf{Layer Type}	&\textbf{Layer Parameters}	&\textbf{Output}	&\textbf{S.C.} \\ \hline \hline

 & & \textbf{Backbone Stage} & & \\ \hline
BB1 &	PointNetSetAbstractionMsg &	$0, [0.1,0.2,0.4], [16,32,128], [[16, 16, 16],[16, 16, 16],[32, 32, 32]]$ &	$(64, 512)$ & \\	
BB2	& PointNetSetAbstractionMsg &	$64, [0.2,0.4,0.6], [32,64,128], [[32, 32, 32],[32, 32, 32],[64, 64, 64]]$	& $(128, 128)$ &	\\
BB3	& PointNetSetAbstractionMsg &	$128, [0.6,0.8,0.9], [64,96,128], [[64, 64, 64],[64, 64, 64],[128, 128, 128]]$	& $(256, 32)$ & \\	
BB4	& PointNetSetAbstraction	& $256, None, None, [256]$ & 	$(256, 1)$	& \\ \hline

 & & \textbf{Feature Cross Referencing Stage} & & \\ \hline
FCR1	&FeatureCrossReferencing	&$256, [128,128]$	    &$(128, 32)$	&    \\
FCR2	&FeatureCrossReferencing	&$128+256, [64, 64]$	&$(64, 128)$	&BB3 \\
FCR3	&FeatureCrossReferencing	&$64+128, [32, 32]$	    &$(32, 512)$	&BB2 \\ \hline
						
 & & \textbf{Feature Re-Encoding Stage} & & \\ \hline
FRE1	&FeatureReEncoding	&$32+64, 0.4, 32, [96, 96]$	        &$(96, 128)$	&BB1 \\
FRE2	&FeatureReEncoding	&$96+64+128, 0.8, 32, [288, 288]$	&$(288, 32)$	&BB2, FCR2 \\
FRE3	&FeatureReEncoding	&$288+128+256, None, None, [672,672]$	&$(672, 1)$	&BB3, FCR1 \\ \hline
						
 & & \textbf{Segmentation Stage} & & \\ \hline
FP1	&PointNetFeaturePropagation	&$672+288, [256, 256]$	&$(256, 32)$	&FRE2 \\
FP2	&PointNetFeaturePropagation	&$256+96, [256, 128]$	&$(128, 128)$	&FRE1 \\
FP3	&PointNetFeaturePropagation	&$128+32, [128, 128]$	&$(128, 512)$	&FCR3 \\
FP4	&PointNetFeaturePropagation	&$128+6, [128, 128]$	&$(128, 1024)$	& \\ \hline

 & & \textbf{Point-wise Classification Stage} & & \\ \hline
FC1	&FullyConnectedLayer	&$128, 128, 0.5$	&$(128, 1024)$ &	\\
FC2	&FullyConnectedLayer	&$128, \#parts$	&$(\#parts, 1024)$ &	\\
\hline					
						
\end{tabular}}
\end{center}
\vspace{-0.4cm}
\caption{Network Architecture of MARNet Part-Segmentation Network. Here, \textit{Output} = (\#output channels, \#output points), \textit{S.C.} = shortcut connections indicating features passed from previous layers.}
\label{tab:MARNetpartseg}
\vspace{-0.4cm}
\end{table*}

\begin{table*}[!ht]
\footnotesize
\begin{center}
{\begin{tabular}{ccccc}
\hline
\textbf{Layer}	&\textbf{Layer Type}	&\textbf{Layer Parameters}	&\textbf{Output}	&\textbf{S.C.} \\ \hline \hline

 & & \textbf{Backbone Stage} & & \\ \hline
BB1	&PointNetSetAbstraction	&$0, 0.2, 32, [32, 32, 32]$	    &$(64, 512)$ \\
BB2	&PointNetSetAbstraction	&$32, 0.4, 32, [64, 64, 64]$     &$(128, 128)$ \\
BB3	&PointNetSetAbstraction	&$64, 0.8, 32, [128, 128, 128]$  &$(256, 32)$ \\
BB4	&PointNetSetAbstraction	&$128, None, None, [256]$	    &$(256, 1)$ \\ \hline

 & & \textbf{Feature Cross Referencing Stage} & & \\ \hline
FCR1	&FeatureCrossReferencing	&$256, [128,128]$	&$(128, 32)$ & \\	
FCR2	&FeatureCrossReferencing	&$128+128, [64, 64]$	&$(64, 128)$ &	BB3 \\
FCR3	&FeatureCrossReferencing	&$64+64, [32, 32]$	&$(32, 512)$	& BB2 \\ \hline

 & & \textbf{Feature Re-Encoding Stage} & & \\ \hline
FRE1	&FeatureReEncoding	&$32+32, 0.4, 32, [64, 64]$	&$(96, 128)$	&BB1 \\
FRE2	&FeatureReEncoding	&$64+64+64, 0.8, 32, [192, 192]$ &$	(288, 32)$	& BB2, FCR2 \\
FRE3	&FeatureReEncoding	&$192+128+128, None, None, [448, 448]$	&$(672, 1)$	& BB3, FCR1 \\ \hline

 & & \textbf{Classification Stage} & & \\ \hline
FC1	&FullyConnectedLayer	&$448, 512, 0.4$	&$(512, 1)$ & \\
FC2	&FullyConnectedLayer	&$512, 256, 0.5$	&$(256, 1)$ & \\
FC3	&FullyConnectedLayer	&$256, \#classes$	&$(\#classes, 1)$ & \\
\hline					
						
\end{tabular}}
\end{center}
\vspace{-0.4cm}
\caption{Network Architecture of MARNet Lite Classification Network. Here, \textit{Output} = (\#output channels, \#output points), \textit{S.C.} = shortcut connections indicating features passed from previous layers.}
\label{tab:MARNetlite}
\vspace{-0.4cm}
\end{table*}

\end{document}